\documentclass[pdflatex,sn-basic]{sn-jnl}

\usepackage{graphicx}%
\usepackage{multirow}%
\usepackage{amsmath,amssymb,amsfonts}%
\usepackage{amsthm}%
\usepackage{mathrsfs}%
\usepackage[title]{appendix}%
\usepackage{xcolor}%
\usepackage{textcomp}%
\usepackage{manyfoot}%
\usepackage{booktabs}%
\usepackage{algorithm}%
\usepackage{algorithmicx}%
\usepackage{algpseudocode}%
\usepackage{listings}%

\theoremstyle{thmstyleone}%
\theoremstyle{thmstyletwo}%

\theoremstyle{thmstylethree}%

\begin{document}

\title[Article Title]{Zero-shot data citation function classification using transformer-based large language models (LLMs)}

\author*[1]{\fnm{Neil} \sur{Byers}}\email{npbyers@lbl.gov}
\author[1]{\fnm{Ali} \sur{Zaidi}}
\author[1]{\fnm{Valerie} \sur{Skye}}
\author[1]{\fnm{Chris} \sur{Beecroft}}
\author[1]{\fnm{Kjiersten} \sur{Fagnan}}

\affil[1]{\orgdiv{DOE Joint Genome Institute}, \orgname{Lawrence Berkeley National Laboratory}, \orgaddress{\street{1 Cyclotron Road}, \city{Berkeley}, \postcode{94720}, \state{California}, \country{USA}}}

\abstract{Efforts have increased in recent years to identify associations between specific datasets and the scientific literature that incorporates them. Knowing that a given publication cites a given dataset, the next logical step is to explore how or why that data was used. Advances in recent years with pretrained, transformer-based large language models (LLMs) offer potential means for scaling the description of data use cases in the published literature. This avoids expensive manual labeling and the development of training datasets for classical machine-learning (ML) systems. In this work we apply an open-source LLM, Llama 3.1-405B, to generate structured data use case labels for publications known to incorporate specific genomic datasets. We also introduce a novel evaluation framework for determining the efficacy of our methods. Our results demonstrate that the stock model can achieve an F1 score of .674 on a zero-shot data citation classification task with no previously defined categories. While promising, our results are qualified by barriers related to data availability, prompt overfitting, computational infrastructure, and the expense required to conduct responsible performance evaluation.}

\keywords{large language model, citation function analysis, natural language processing, machine learning}

\maketitle

\section{Introduction}\label{sec1}
The past two decades have seen a surge of interest regarding the identification of data citations in the scientific literature. By "data citation", we refer to any mention, either formal or informal, of specific data or datasets in a given publication. Organizations like CrossRef \citep{hendricks_crossref_2020}, DataCite \citep{neumann_datacite_2014}, and Make Data Count \citep{cousijn_bringing_2019} have made strides in building the infrastructure required to make data citations machine readable and resolvable. The ICPSR Bibliography of Data-related Publications \citep{icpsr_bib_2025}, and more recently, these authors' own work on the Data Citation Explorer \citep{byers_identifying_2024} illustrate several attempts to maintain registries of domain-specific data citations by data generators themselves.

Identifying which publications cite which data is an important first step. The next step necessarily involves investigation of the purposes for which authors cite certain data. Literature on this task, variously referred to as citation "function" analysis or citation "intent" analysis, has traditionally focused on identifying authors' purposes for citing other literature \citep{teufel_automatic_2006,roman_citation_2021,zhang_deep_2025}. Data can be cited for any variety of reasons, including, but not limited to, reports on the generation and availability of new datasets, inclusion of existing public datasets in large-scale aggregate analyses, testing and benchmarking new analytical software, training machine-learning models, and providing illustrative examples for specific methods or protocols. The distribution of such use cases for a given dataset or corpus holds important implications for a variety of stakeholders, including data generators, potential future users of that data, infrastructure providers, and funding organizations. For example, scientists might use this information to brainstorm novel ways for looking at old data. Repositories and other providers might use this information to guide development of tools and infrastructure that can best support ongoing research trends, and so on.

Manual evaluation of data citations and the development of training datasets for classical machine-learning algorithms are both slow and resource intensive, limiting their usefulness as scalable solutions \citep{zhang_hybrid_2023,chatrinan_overcoming_2025}. Because they are pretrained, LLMs offer promise as zero-shot classifiers that can avoid the expense associated with manual annotation or training set development. Others have begun to explore the application of language models and deep learning more generally to citation function classification tasks, though notably with a distinct focus on citations of scientific research articles as opposed to citations of data \citep{jiang_contextualised_2023,budi_understanding_2022,liu_low-resource_2023,c_citation_2024,cohan_structural_2019,nambanoor_language_2024}.

While many previous studies focus on the immediate semantic context surrounding the citation of a given research artifact within a scientific publication, others have noted that more information is necessary to gain a comprehensive understanding of a given citation \citep{zhang_towards_2022}. This is particularly true for publications that cite data, where clues concerning data usage are often sprinkled throughout multiple sections. With the growth of context window sizes in the past few years \citep{torene_window_2025}, LLMs also offer the benefit of accepting large blocks of text as inputs. This feature enables the consumption of entire scientific publications and all of the implicit, cross-document semantic relationships they contain.

Here we report the application of an "off-the-shelf" LLM, Llama 3.1-405B, to a complex data citation classification task. Given a set of previously identified pairs between publications and specific accession numbers for genomic data, we ask the model to describe various aspects of the authors' usage of these data. This task is difficult because each document can have multiple labels and because no discrete set of known labels exists. This is because we do not as of yet have an authoritative schema documenting all of the ways that data can be used, nor have we documented all of the tools that can be used with data as part of this usage. The lack of a defined label set in particular contrasts with many recent citation function classification studies attempting to produce labels in alignment with proposed schema \citep{jha_nlp-driven_2016,hernandez-alvarez_citation_2017,dong_ensemble-style_2011,jurgens_measuring_2018,su_neural_2019}. A further layer of difficulty is introduced by the zero-shot nature of the task, meaning that no training examples are provided to adjust model parameters prior to inference. We approach the problem using a decision-tree prompt structure with several Retrieval-Augmented-Generation (RAG) components \citep{lewis2021retrievalaugment}. We refer in the present work to this larger classification workflow as a "machine assistant".

The ability of LLMs to provide reliable, structured outputs as part of complex tasks has not been consistently or comprehensively evaluated. With applications like those mentioned above, for example decision-making regarding scientific infrastructure development, strict quality control standards must exist when exploring the use of LLMs. In pursuance of this consideration, we also report an evaluation framework called SARGO (Selective Aggregation of Redundant and Granular Outputs) for determining the efficacy of our machine assistant. The framework allows us both to more confidently benchmark model performance on a specific task and to discuss said performance using metrics (precision, recall, F1) allowing easier comparisons between ours and other retrieval methods.

We find that while our machine assistant demonstrates promising performance on a difficult, zero-shot, domain-specific task, that performance does not currently meet requirements for robust research assessment applications. The significant resource requirements for building and confidently evaluating an LLM-based system for data citation classification are discussed, as are LLM-specific limitations inhibiting systematic refinement of our machine assistant's workflows. We argue that these requirements and limitations still present significant barriers for widespread adoption of LLM-based systems for scalable data citation classification. 

\section{Methods}\label{sec2}

\subsection{Data Acquisition: The Data Citation Explorer}\label{sec2a}

The data citation corpus used in this analysis was produced by the Data Citation Explorer (DCE), described in detail elsewhere \citep{byers_identifying_2024}. Briefly, the DCE builds a registry of certain classes of identifiers stored metadata systems at the US DOE Joint Genome Institute (JGI) that refer to genomic datasets either produced by JGI or hosted by JGI's suite of online data resources \citep{chen_imgm_2022,goodstein_phytozome_2011,grigoriev_mycocosm_2013,grigoriev_phycocosm_2020,jgi_portal_2025}. These identifiers are then used to crawl available full-text literature for mentions of the datasets they represent. The DCE then stores any hits as structured, queryable linkages between datasets and citing publications. At the time of capture (November 11, 2024), the resulting corpus consisted of pairs between 43,620 unique publications and 51,961 unique identifiers. Of this initial set, some 38,053 publications were retrieved by the DCE as a result of full-text hits. 35,216 of these were available for full-text analysis via PubMed Central. These publications were linked to 49,833 identifiers.

\subsection{Text Preprocessing}\label{sec2b}

To facilitate downstream text analysis, we developed a set of parameterized XSLT 1.0 stylesheets (JATS v1.2 XML, JATS v1.3 XML) to transform PubMed Central (PMC) XML records into plain text conforming to the W3C Text technique T3 \citep{w3_t3_2025}. These transformations were accomplished using a Java 8 client library that interfaces directly with both the PubMed Central (PMC) and PMC-OAI APIs. The library enables retrieval and conversion of publications within data processing pipelines and incorporates error handling and rate-limiting mechanisms to ensure reliable operation against NCBI API endpoints.

Parameters were set to include or exclude certain components of each publication. For our experiments, we chose to include tables and all text content associated with figures. Front and back matter (e.g. author information, references, and supplemental file descriptions) were removed, as were section headers. The stylesheets support an optional chunking mode, which partitions the text into smaller, semantically coherent blocks suitable for processing with large language models (LLMs) operating with limited context windows. This feature was not used to generate the reported results for reasons discussed below.

\subsection{Prompting and Retrieval-Augmented Generation (RAG)}\label{sec2c}

For each publication-identifier pair, the LLM was prompted to generate outputs answering the following questions:

\begin{enumerate}[1.]
\item Data Accessed (True/False): Was the data represented by the identifier accessed and used in the analyses presented in the linked publication?

\item Use Cases (Array): What distinct use cases for the data represented by the identifier were described by the linked publication?

\item Software/Tools (Array): What software or analytical tools were used with the data represented by the identifier as part of the listed use cases?
\end{enumerate}

\begin{figure}[ht]
\centering
\includegraphics[width=0.7\textwidth]{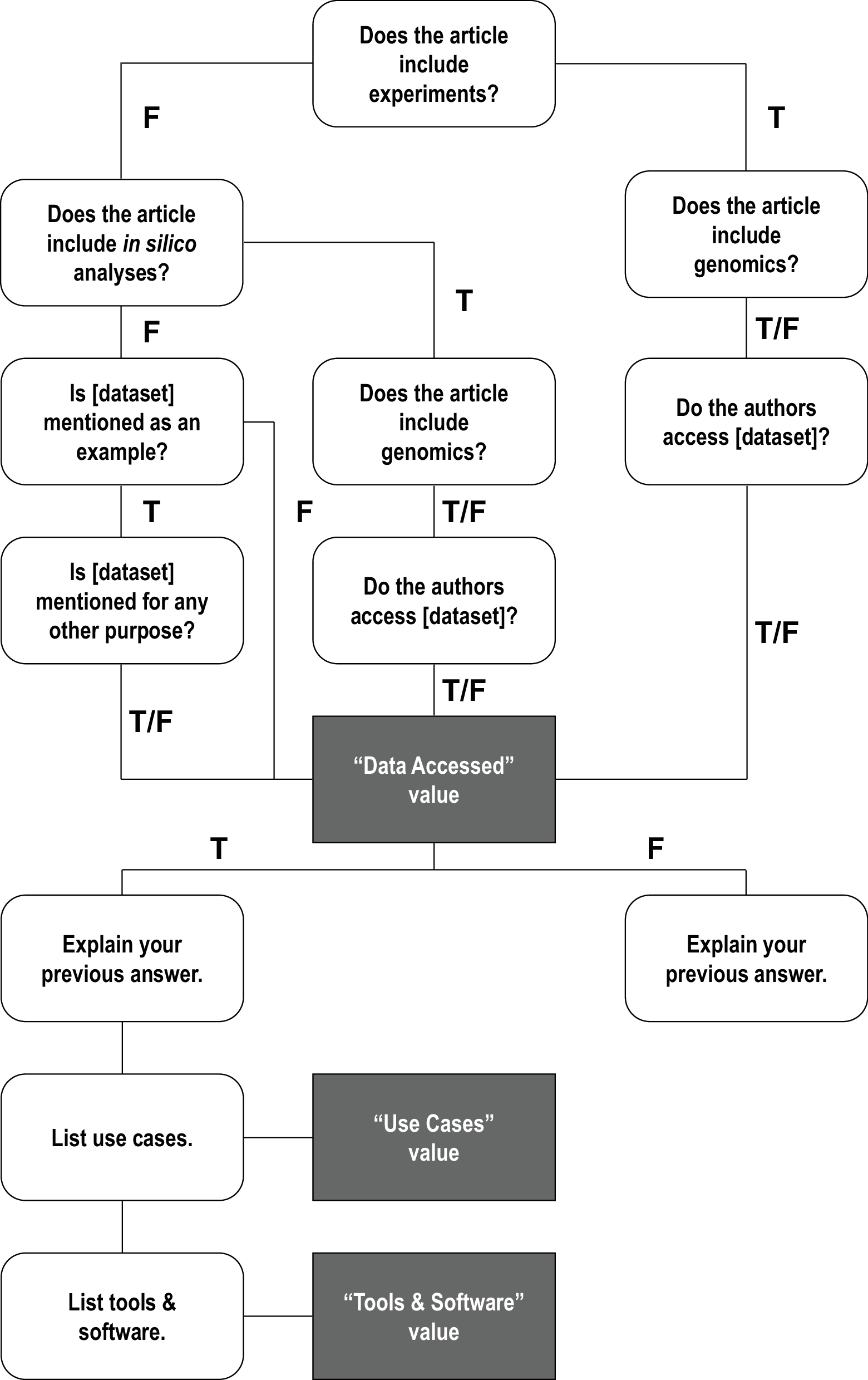}
\caption{Decision tree prompt structure underlying the machine assistant's workflow. The final outputs for a given publication-accession pair are represented by the dark gray boxes.}\label{fig1}
\end{figure}

This was achieved with a decision tree prompting structure. In this structure, each identifier-publication pair was subjected to its own "chat" consisting of multiple successive prompts. The final outputs of each chat are those representing the required values in the above list. Briefly, the model is first provided with the full-text of the publication in question. The model is then prompted to discern whether or not the publication contains experiments or in silico bioinformatics analyses. Based on answers to these questions, the model is then prompted to explain any mentions of the identifier in question. If it is determined that the authors accessed the data in question, the model is then prompted to provide information on tools and use cases involving that data. The decision tree is illustrated in Figure \ref{fig1}. Verbatim prompt language can be found at \url{https://github.com/npbyers0401/citation-function-analysis}. Refer to Tables ~\ref{tab1} and \ref{tab2} for example outputs for each category.

Retrieval-Augmented Generation (RAG) is implemented in the workflow to provide the model with context about the linked identifiers using information from NCBI and JGI metadata systems. Statements containing information about a provided identifier are constructed by the DCE according to business logic that takes into account the identifier's class and available linked metadata. The included information was selected based on assumptions about which metadata types might provide the most helpful guidance in the context of most scientific publications. Two examples for different identifier classes can be seen below. Extracted metadata are highlighted in bold:

\begin{enumerate}[1.]
\item CP001672 - The accession "CP001672.1" refers to a \textbf{Nucleotide Sequence} record from the \textbf{National Center for Biotechnical Information (NCBI) GenBank Nucleotide Database}. The record contains \textbf{nucleotide sequence data} from \textbf{Methylotenera mobilis JLW8} (a \textbf{Prokaryote} from the species \textbf{Methylotenera mobilis}). The sequenced strain is referred to as "\textbf{JLW8}".

\item GCF\_000696285 - The accession "GCF\_000696285.1" refers to an \textbf{Assembly} record from the \textbf{National Center for Biotechnical Information (NCBI) Assembly Database}. The record contains \textbf{information about a Genome Assembly} from a biological sample identified as \textbf{SAMN02732238}. The sample is from \textbf{Bacillus amyloliquefaciens} (a \textbf{Prokaryote} from the genus \textbf{Bacillus}).
\end{enumerate}

\subsection{Hardware, Parameters, and Environment}\label{sec2d}

The machine assistant workflow was implemented in Python. As stated before, we selected Llama 3.1-405B \citep{grattafiori_llama_2024} as the base model. This model was selected because it was the largest open-weight model available at the time we conducted our experiments. For our purposes, it is essential that any model used be open-weight to ensure that results can be reproduced locally even after a model has been sunsetted. Hugging Face Transformers v4.50.0 was used for porting the version of Llama 3.1-405B used in our experiments. The model was quantized to 4-bit precision using bitsandbyes v0.45.5 in order to maximize memory efficiency. Other important packages included the OpenAI Python API library (v.1.76.2), PyTorch (v2.6.0+cu118), Hugging Face Accelerate (v1.6.0), and Tokenizers (v0.21.1). Computation was accomplished using single nodes on Berkeley Lab's Lawrencium cluster equipped with either four Nvidia A-100 or eight Nvidia H-100 GPUs. Our quantized version of the model required some 200+ GB of GPU RAM for the model weights alone, with additional RAM requirements for each forward pass (ex. for utilizing a key-value cache storing intermediate values). Sampling was disabled for each pass to ensure the generation of deterministic, reproducible results.

\subsection{Sampling and Manual Annotation}\label{sec2e}

Two samples of publications were selected for developing and evaluating our machine assistant's workflow, respectively. The first set, which we refer to as the "initial" set, was selected in a semi-random fashion. Five publications were chosen at random with no preference for specific criteria. In order to gather publications of various ages and lengths, eight publications were selected by choosing one publication from each quartile of both the publication year and raw character length fields. Following that, ten further publications were selected by manually choosing publications from distinct publishers and that cite diverse identifier classes. This was done to avoid overrepresentation in both of these fields. One publication was later excluded because it was found to have been included in the DCE corpus as a result of a technical error. The initial set consisted of 22 publications linked to 56 unique identifiers and was used for iterative testing and development of our workflow.

20 further publications were sampled without replacement to create an "evaluation" set. These publications were chosen according to a more rigorous stratified random sampling method than that used to select the initial set. Publications included in the evaluation set were split between publications published before and after the training of the model used in the experiments (Llama 3.1-405B, December 2023), with the justification that publications published before the model's training could conceivably have been included in the training dataset and thus represented in the model's parameters. Each resulting subset was then sampled randomly from strata representing diverse publishers, publication ages, character counts, and cited identifier classes. Two publications previously identified as false positive associations made by the DCE (publications in which the data represented by the mentioned accession was not actually used in the publication) were added to the evaluation set to determine the model's ability to identify such cases. The evaluation set consisted of 20 publications linked to 34 unique identifiers and was used only in the final evaluation of our workflow.

All publication-identifier pairs in each sample set were manually annotated. For each pair, two human domain experts independently assigned values for the three fields that constituted the model outputs: Data Accessed, Use Cases, and Software/Tools. Both experts then met to resolve conflicts and generate consensus assignments. The resulting annotations were used as the "gold standard" for evaluating model outputs. Machine outputs were viewed and evaluated only after consensus versions of the gold standard set were created to avoid influencing the expert reviewers. All publications, linked identifiers, and manual annotations used in this analysis can be found at \url{https://doi.org/10.5281/zenodo.17401492}.

\subsection{Evaluation and Scoring}\label{sec2f}

Precision, Recall, and F1 score are common metrics used to evaluate classifiers and information retrieval systems. To evaluate the outputs generated by the machine assistant, we compared them with our gold standard and determined both relevancy (alignment between human and machine) and retrieval (proportion of human values retrieved) rates. Doing so allowed us to calculate precision, recall, and F1 values and contextualize our machine assistant's performance in a more interpretable and familiar context than most LLM benchmarks.

For those fields without known class lists ("Tools / Software" and "Use Cases"), values included in both gold standard and machine output sets were considered true positives. Values included in the gold standard set but not the machine set were considered false negatives. Conversely, values included in the machine set but not the gold standard set were considered false positives. True negative values were not calculated for these fields. This is because the total number of conceivable values for each of these fields is likely infinite or not practically knowable, and in any case this statistic is not used when calculating our chosen metrics. Assignments were made differently for the binary "Data Accessed" field. "TRUE" values for this field in the machine-generated set were marked as true positives if also marked as "TRUE" by the human expert. "FALSE" values in the gold standard set that were marked as such by the machine assistant were marked as true negatives, and so on.

Granularity issues that characterize our problem statement made a simple application of these metrics difficult. How one describes the way in which a dataset is used or mentioned can be high-level and basic, specific and detailed, or anywhere in between. For example, one could describe a publication \citep{tsolis_genome_2009} as including a "comparative analysis" between genomes. One could also describe that same publication as including "comparative analysis of pseudogene content", "identification of shared or unique protein-coding genes",  and "comparative best-match BLASTp searches", all of which constitute components of the larger "comparative analysis". Where one lands on this granularity spectrum is arbitrary, and articulating just how granular a requested set of descriptions should be often leads to inconsistent results. This issue led to many situations in which the number of statements generated to describe singular use cases differed between our human and machine evaluators, thus rendering the tabulation of false positives, true positives, etc. impossible. This is because recall assumes that all elements in a queried corpus are discrete and unique. One cannot have multiple retrieved results that correspond to a single relevant element from the original corpus, because such a scenario could lead to situations in which the numerator (relevant retrieved results) is greater than the denominator (all relevant elements).

Redundancy in outputs generated by machine assistants compounded the granularity issue. We noticed in testing that our machine assistant often generated several very closely related or analogous descriptions related to a single use case mentioned in a publication (again violating the one-to-one requirement for calculating recall). For example, one publication in our initial set \citep{frisch_invasive_2018} refers to a genome being used as the "root" and also as the "outgroup" in the construction of a phylogenetic tree. These two descriptive words describe a single analytical operation run by the authors, something a human domain expert would recognize quickly. However, the difference in vocabulary ("root" vs. "outgroup") that appears with each dataset mention in the publication itself led our machine assistant to generate two use cases for this operation. Such behavior, to which we refer as "spitballing", by our machine assistant occasionally led to an inflation of redundant statements describing singular instances of use and added further difficulty to our evaluation metric calculations.

We introduced a simple mechanism, which we coin here as SARGO (Selective Aggregation of Redundant and Granular Outputs), to accommodate for both issues. Simply put, discrete but redundant or high-granularity machine outputs that corresponded to a smaller number of human-generated values were aggregated to correspond in number to their human-generated counterparts. Using our "root" and "outgroup" redundancy example from above, both of these machine outputs were aggregated as a single output and marked as a single true positive result when compared to the human-generated value for the same use case.

The same process also accommodates for the granularity issue. For example, our set included several publications for which the machine assistant generated multiple high-granularity descriptions of subprocesses involved with the production of a published genome. The human evaluators grouped these subprocesses, including sequencing, annotation, and deposition of the genome into a single value. SARGO was applied here to aggregate and mark the machine outputs as a single true positive result. Because the prompt language explicitly requests that the machine assistant generate high-granularity responses, in only one instance was SARGO applied backwards by aggregating multiple human-generated responses to match a single machine-generated output. An example of a confusion matrix resulting from a single publication's evaluation with SARGO applied to aggregate several granular use cases is provided in Table \ref{tab1}. A similar example with SARGO applied to aggregate two redundant use cases is proved in Table \ref{tab2}. 

\begin{table}[ht]
\begin{tabular}{ p{2cm}p{3cm}p{3cm}p{2cm} } 
\toprule
\multicolumn{4}{c}{\large{Evaluation Matrix}} \\ \cmidrule(lr){2-3}
\textbf{Category} & \textbf{Human Value} & \textbf{Machine Value} & \textbf{Assessment}\\ 
\midrule
Data Accessed & TRUE & TRUE & TP\\
\midrule
\multirow{9}{2cm}{Tools and Software} & BEAST 2.4.5 & & FN \\ \cmidrule(lr){2-4}
& ClonalFrameML & & FN \\ \cmidrule(lr){2-4}
& gingr & & FN \\ \cmidrule(lr){2-4}
& Interactive Tree of Life & & FN \\ \cmidrule(lr){2-4}
& Parsnp & & FN \\ \cmidrule(lr){2-4}
& PhyML & PhyML & TP \\ \cmidrule(lr){2-4}
& R & & FN \\ \cmidrule(lr){2-4}
& RAxML 8.2.11 & RAxML & TP \\ \cmidrule(lr){2-4}
& VariScan & & FN \\
\midrule
\multirow{3}{2cm}{Use Cases} & Outgroup in a molecular dating analysis & & FN \\  \cmidrule(lr){2-4}
& \multirow{2}{3cm}{Outgroup in a phylogenetic analysis} & Outgroup selection in phylogenetic analysis & TP \\   \cmidrule(lr){3-4}
& & Rooting of the phylogenetic tree & Aggregated \\
\midrule
\midrule
& \multicolumn{3}{c}{\large{Confusion Matrix}} \\ \cmidrule(lr){2-4}
& & \textbf{Retrieved} & \textbf{Not Retrieved}\\ \cmidrule(lr){3-4}
& \textbf{Relevant} & 4 & 8\\ \cmidrule(lr){3-4}
& \textbf{Irrelevant} & 0 & 0\\
\bottomrule
\end{tabular}
\caption{Evaluation matrix for \cite{frisch_invasive_2018} containing multiple redundant machine-generated use case descriptions and resulting confusion matrix. The cited accession is \href{https://www.ncbi.nlm.nih.gov/nuccore/CP000046}{CP000046.1}.}\label{tab1}
\end{table}

\begin{table}[ht]
\begin{tabular}{ p{2cm}p{3cm}p{3cm}p{2cm} } 
\toprule
\multicolumn{4}{c}{\large{Evaluation Matrix}} \\ \cmidrule(lr){2-3}
\textbf{Category} & \textbf{Human Value} & \textbf{Machine Value} & \textbf{Assessment}\\ 
\midrule
Data Accessed & TRUE & TRUE & TP\\
\midrule
\multirow{4}{2cm}{Tools and Software} & COG & & FN \\ \cmidrule(lr){2-4}
& BLAST & BLAST & TP \\ \cmidrule(lr){2-4}
& MUMmer & MUMmer & TP \\ \cmidrule(lr){2-4}
& & GLIMMER & FP \\
\midrule
\multirow{3}{2cm}{Use Cases} & \multirow{2}{3cm}{Used in a comparative study of Brucella genomes to investigate veterinary pathogenicity} & comparative analysis of genomic features & TP \\  \cmidrule(lr){3-4}
& & comparative best-match blastp searches & Aggregated \\   \cmidrule(lr){3-4}
& & suffix tree analysis using MUMmer & Aggregated \\
\midrule
\midrule
& \multicolumn{3}{c}{\large{Confusion Matrix}} \\ \cmidrule(lr){2-4}
& & \textbf{Retrieved} & \textbf{Not Retrieved}\\ \cmidrule(lr){3-4}
& \textbf{Relevant} & 4 & 1\\ \cmidrule(lr){3-4}
& \textbf{Irrelevant} & 1 & 0\\
\bottomrule
\end{tabular}
\caption{Evaluation matrix for \cite{tsolis_genome_2009} containing multiple overly granular machine-generated use case descriptions and resulting confusion matrix. The cited accession is \href{https://www.ncbi.nlm.nih.gov/nuccore/NC_003317}{NC\_003317}.}\label{tab2}
\end{table}

Evaluation was performed in a semi-automated manner. Machine-generated outputs that could be linked to human-generated outputs via simple string matches were marked as true positives. Those that could not be matched in this way were manually reviewed. SARGO was applied manually as well, with a human expert making decisions on how to aggregate redundant or overly granular machine outputs.

\section{Results}\label{sec3}

The full set of precision, recall, and F1 scores for our machine assistant can be seen in Table \ref{tab3}. Metrics were calculated twice: once for each full run of machine-generated outputs based on the initial and evaluation sets. Metric values for each run can be further broken down by the response categories "Data Accessed", "Use Cases", and "Tools and Software." Our overall F1 scores were .805 and .674 for the initial and evaluation sets, respectively. Scores were generally higher for the "Data Accessed" category than for the other two categories, though this is likely due in part to the differences in difficulty between binary classification tasks and tasks with unknown label sets. These results indicate that the machine assistant was effective in determining whether or not authors actually accessed a dataset for tangible use, though performance degraded when the assistant was asked to generate specific information about that usage.

\begin{table}[ht]
\begin{tabular}{ cccccc } 
\toprule
\textbf{Set} & \textbf{Model} & \textbf{Value Type} & \textbf{Recall} & \textbf{Precision} & \textbf{F1}\\ 
\midrule
\midrule
\multirow{4}{*}{Initial} & \multirow{4}{*}{Llama 3.1-405B} & Data Accessed & 1 & 1 & 1 \\ \cmidrule(lr){3-6}
&  & Use Cases & .822 & .845 & .833 \\ \cmidrule(lr){3-6}
&  & Tools and Software & .686 & .813 & .744 \\ \cmidrule(lr){3-6}
&  & \textbf{Overall} & \textbf{.76} & \textbf{.851} & \textbf{.805} \\
\midrule
\midrule
\multirow{4}{*}{Evaluation} & \multirow{4}{*}{Llama 3.1-405B} & Data Accessed & 1 & .964 & .982 \\ \cmidrule(lr){3-6}
&  & Use Cases & .653 & .8 & .719 \\ \cmidrule(lr){3-6}
&  & Tools and Software & .476 & .758 & .585 \\ \cmidrule(lr){3-6}
&  & \textbf{Overall} & \textbf{.579} & \textbf{.805} & \textbf{.674} \\
\bottomrule
\end{tabular}
\caption{Evaluation metrics for both the initial and evaluation publication sets.}\label{tab3}
\end{table}

The most salient outcome lies in the difference in scores between our initial and evaluation sets. The overall F1 score was .131 greater for our initial set as compared with the results for our evaluation set. We iteratively and manually tuned our prompt structure and model parameters to achieve the best possible results on the initial set. We then used the resulting structure to generate results for the evaluation set, which was purposefully not used for any tuning. This allowed us to determine the extent to which our prompts and parameters were "overfit" to our initial set by determining their performance with unseen examples. These results show significant drops in performance comparing the evaluation set outcomes with the initial set outcomes. The performance drops are more salient in the undefined "Tools and Software" and "Use Cases" categories than they are in the binary "Data Accessed" category. The degree of the overall differences in performance between the two sets illustrates the "brittleness" of our prompt structure and speaks to the limits of prompt engineering as a standalone solution for domain-specific LLM tasks.

For most tasks our machine assistant achieved higher precision than recall. Though the system failed to retrieve a significant number of relevant results, those that were returned were of comparably high relevance. We also observed a smaller drop in overall precision (.046) than for recall (.181) between the initial and evaluation sets, indicating that the prompt "brittleness" discussed above degrades our assistant's ability to retrieve relevant results more than it encourages the assistant to hallucinate false positive results.

We also note that the values in the "Tools and Software" category for a given paper are heavily dependent on the values in the "Use Cases" category. There are often one-to-many relationships between single analytical processes in publications and the tools that are required to accomplish them, while the reverse is less often true. The practical effect of this on our assistant's performance is that failing to retrieve a single use case will likely cause a failure to retrieve one or more relevant values for the "Tools and Software" category. For example, should the assistant fail to identify that a genome was used in a phylogenetic analysis \citep{frisch_invasive_2018}, it would also fail to retrieve the multitude of software tools that were part of that analysis. Cases also occurred in which the model would correctly identify a given use case, but would fail to retrieve one or more associated "Tools and Software" values from the publication.

\section{Discussion}\label{sec4}

\subsection{Assessment in Context}\label{sec3a}

Our machine assistant's ability to achieve an F1 score of .674 on a difficult, complex, domain-specific, zero-shot classification task with no defined labels indicates promise for the application of LLMs to tasks previously thought outside the capabilities of ML-based systems. This promise must be qualified, however, by the recognition that such performance remains below the standards necessary for robust research assessment activities. To relate this performance back to concepts more often associated with LLMs, one can use precision to determine what might be considered a "hallucination rate" ($1-P$). For our evaluation set we achieved an overall precision value of .805. The hallucination rate would thus be .195, meaning that around 20\% of results generated by the system should be labeled as "hallucinations." Further, the recall values in our experiment can be seen as an indicator of how well our system performs a "Needle-in-a-Haystack" task, with higher recall values indicating greater ability for "finding" the needles. With an overall recall value of .582 for our evaluation set, we could expect our system to miss 40-50\% of the needles in our haystack. Application of a system with such caveats to a task like the systematic evaluation of JGI's product offerings, for example, would likely lead to a greatly misleading overall picture of our user community's activities. What performance might be considered "robust" will vary between contexts. For the proposed research assessment applications at JGI, we would strive for an F1 score of at least .95 before considering a production implementation of our workflow.

This study underscores the importance of evaluating any and all LLM-generated outputs that are part of workflows intended to produce robust, actionable information. That our machine assistant both failed to retrieve a considerable portion of extant relevant results and also generated a significant number of hallucinated responses illustrates the caution with which we should approach the application of LLMs to even focused, localized tasks. This consideration becomes even more important in "agentic" workflows where LLMs might be asked to both generate individual outputs and make decisions regarding the synthesis of outputs from like systems. Many benchmarks exist for gauging the performance of LLMs on difficult tasks \citep{clark_think_2018,guha_legalbench_2023, xie_finben_2024, singhal_large_2023, wang_mmlu-pro_2024, phan_humanitys_2025, rein_gpqa_2023}, but these tests often suffer from having known structures/responses that can be gamed. The implications of this are demonstrated by our machine assistant's significantly improved performance on known examples (the initial set) as compared with unseen examples (the evaluation set).

\subsection{Future Directions}\label{sec3b}

"Fine-tuning" (or "post-training") offers potential means for improving our results. This is achieved by leveraging a set of pre-labeled examples to modify a pretrained model's weights so that it might be better suited to the task at hand. We argue that while we may achieve better scores with a fine-tuned model than with a stock model, the expense in creating the labeled dataset required for fine-tuning would effectively eliminate the "zero-shot" component of our system and defeat a primary motivation for applying LLMs to our specific problem. The complexity of our task makes it likely that we would need many more examples in addition to the initial and evaluation sets we already hand-labeled. A new set would also need to be created every time a model is deployed in a new domain context. Each publication evaluated in this study took human experts anywhere from 30 to 60 minutes to evaluate individually, with an additional 30-60 minutes per publication for the experts to consolidate their annotations. Doing so for the hundreds or thousands of publications required for fine-tuning a model were beyond resources available for this experiment. The issue is further compounded by the necessity of hand-labeling new evaluation datasets periodically so that model performance can be continually assessed over time.

"Chunking" each publication into smaller pieces could improve our results by lowering the overall context size of each pass \citep{hong_context_2025}. Such a method simplifies the "Needle-in-a-Haystack" component of our task by effectively providing the model with a much smaller haystack each time it is called. We avoided this method because many of the examples in our initial and evaluation sets included publications where clues concerning usage of linked datasets are sprinkled throughout the text. Because many LLMs (including our chosen model) are stateless, information contained in one chunk would be lost when asking questions concerning another chunk. We do not as yet have an automated means for determining which chunks of a given publication would be necessary to capture all relevant usage information for inclusion in a single pass, though establishing such a method would constitute grounds for future exploration.

Our results demonstrate that the prompt language carefully crafted for our initial set ended up being relatively poorly suited for the examples in our evaluation set. Natural language is inherently ambiguous, and that ambiguity is at odds with the expectation of deterministic results usually required for generating reliable, structured data and metadata. We worked iteratively to engineer a system of prompts that would not only achieve high scores with the examples at hand, but that would also hopefully generalize well to unseen examples. Often this involved minor tweaks in vocabulary or word order in individual prompts so as to correct an error seen in testing. We noticed that while a single tweak might correct one error in the corpus, it might also end up creating other errors elsewhere in the corpus. One is left in these instances blindly searching to understand the model's embedded connotations for specific terms so as to best guide it to the desired results. Because prompts are created from natural language by hand, it is difficult to hypertune them for optimized results as is possible when working with more classical architectures. Recent innovations in automated prompt development, such as GEPA \citep{agrawal2025gepareflectivepromptevolution} or MIPRO \citep{opsahlong2024optimizing}, could help mitigate such difficulties, and the application of these methods to our workflow are grounds for future experiments.

Being over two years old as of this writing, Llama 3.1-405B has been superseded in most benchmarks by more current models. For example, it scores almost 30 percentage points beneath (\url{https://www.vellum.ai/open-llm-leaderboard}) the most recent models on one common benchmark for complex reasoning, GPQA Diamond \citep{rein_gpqa_2023}. We ran our workflow against our evaluation set using a more recent commercial model, Anthropic's Claude 4.1 Opus \citep{anthropic_claude_2025}, to determine how our results might be affected by model choice.

\begin{table}[ht]
\begin{tabular}{ cccccc } 
\toprule
\textbf{Set} & \textbf{Model} & \textbf{Value Type} & \textbf{Recall} & \textbf{Precision} & \textbf{F1}\\ 
\midrule
\midrule
\multirow{4}{*}{Evaluation} & \multirow{4}{*}{Claude Opus 4.1} & Data Accessed & .926 & 1 & .962 \\ \cmidrule(lr){3-6}
&  & Use Cases & .837 & .82 & .828 \\ \cmidrule(lr){3-6}
&  & Tools and Software & .759 & .714 & .736 \\ \cmidrule(lr){3-6}
&  & \textbf{Overall} & \textbf{.796} & \textbf{.769} & \textbf{.782} \\
\midrule
\midrule
\multirow{4}{*}{Evaluation} & \multirow{4}{*}{Llama 3.1-405B} & Data Accessed & 1 & .964 & .982 \\ \cmidrule(lr){3-6}
&  & Use Cases & .653 & .8 & .719 \\ \cmidrule(lr){3-6}
&  & Tools and Software & .476 & .758 & .585 \\ \cmidrule(lr){3-6}
&  & \textbf{Overall} & \textbf{.579} & \textbf{.805} & \textbf{.674} \\
\bottomrule
\end{tabular}
\caption{A comparison between evaluation metrics for machine-generated values using either Claude Opus 4.1 or Llama 3.1-405B as the base model.}\label{tab4}
\end{table}

These results can be seen in Table \ref{tab4}. The major difference between these results and those we achieved with the older model is the vastly improved recall for the "Tools \& Software" and "Use Cases" fields. Claude 4.1 showed slightly increased precision compared with Llama 3.1 in the "Use Cases" category, with slightly decreased precision in the "Tools and Software" category. In other words, the newer model is much better at the "Needle in a Haystack" component of our task, but suffers from a similar hallucination rate. Though it retrieves a larger portion of relevant results (75-85\% as opposed to 45-65\% with Llama 3.1), an assistant leveraging Claude Opus 4.1 still does not meet standards for robust research assessment. This is especially true in tasks where precision might be the more important metric, or where certain important but low-frequency class types are among those missed by the classifier. To provide a practical example, it would be difficult to justify strategic plans for a government user facility based on data that both excludes 15-20\% of relevant examples and contains hallucinations at a rate of 20-30\%.

\subsection{Resource Requirements}\label{sec3c}

While theoretically offering potential to offload some of the manual costs associated with scaling our citation classification task, an LLM-based system comes with significant resource requirements of its own. Our experiments leveraged existing state-of-the-art high-performance-computing (HPC) resources housed at a DOE National Laboratory, the likes of which would be unavailable to most organizations. The GPU requirements in particular created bottlenecks for our workflows, and even with four or eight 80GB cards we were unable to instantiate an unquantized, full-precision version of Llama 3.1-405B due to exceeded GPU memory. Spreading the load across multiple GPUs created other difficulties associated with data movement between the cards. Either way, such hardware is likely unavailable to most organizations.

There are many commercial providers that allow API access to remotely instantiated models that could be used as a substitute for the locally instantiated model used in our experiments. These providers often charge for usage by the token. Our system requires not only high token-counts per message to accommodate for the full text of scientific publications, but also requires multiple passes with similar-length messages as the machine assistant progresses through our prompt decision tree. This workflow causes token counts to inflate quickly, and with them the potential costs of using a commercial model for our task. At the time of our experiments, the full DCE corpus consisted of 122,292 unique identifier-publication pairs. Applying the tiktoken (\url{https://github.com/openai/tiktoken}) tokenizer to the chats that resulted from the original run with our initial annotated set, we found that the median token counts per pair were 54,600 input tokens and 246 output tokens. Applying current (as of 8/15/2025 - \url{https://aws.amazon.com/bedrock/pricing/}) pricing models a commercial version of Llama 3.1-405B (\$0.0024 per 1000 output/input tokens), this indicates that a median price for generating outputs for the entire DCE corpus would be \$16,097 in API calls alone. The higher cost of using newer commercial models is not insignificant. Using these same calculations, we estimate that a median price for using the version of Claude Opus 4.1 hosted by Google (\$15 per 1m input tokens, \$75 per 1m output tokens) with the entire DCE corpus would be over \$100,000 in API calls alone (as of 10/14/2025 - \url{https://cloud.google.com/vertex-ai/generative-ai/pricing#claude-models}).

Using either local or commercial models in any case would not mitigate costs associated with developing the other components of the system (prompting, RAG, data retrieval, etc.). For example, the DCE is a bespoke system developed and maintained in-house for JGI needs. Many organizations do not have the ability to create a similar system, meaning that the dataset-publication pairs used as inputs in our experiments would not be available to some hoping to recreate this experiment in their own organizational context. This problem is compounded by the gaps in open access (OA), full-text publication data availability. Of the 38,053 publications in our corpus with full-text hits, some 35,216 (92.5\%) were retrievable for natural language processing (NLP) analyses. The same may not hold true for other domains with lower representation in OA publication corpora. To provide a sense of scale, open collections like PubMed Central's OA subset \citep{sayers_database_2024} and the Semantic Scholar's Open Research Corpus \citep{lo-etal-2020-s2orc} index 10-20 million total publications, while paid services like Dimensions index the full-text content of well over 100-million publications \citep{ds_dimensions_2025}. This indicates that comprehensive citation function analysis could be limited by available data.

Finally, SARGO provides us a means for applying common metrics to outputs for our specific task, but other LLM-based classifiers in other contexts will likely require different or additional mechanisms or scoring procedures. The production of our two manually-annoted sets would also need to be replicated for other tasks, or for a similar task in a different organizational or domain context. Planning and performing evaluation work is expensive. Though unglamorous, it requires as much creativity as the development of analysis workflows themselves. As our results indicate, a robust evaluation component should be considered a strict prerequisite for the adoption of LLMs in scientific settings and should be included in the budget for any analogous projects.

\section{Data Availability}\label{sec5}

All data used in the presented analyses can be found at the following Zenodo repository: \url{https://doi.org/10.5281/zenodo.17401492}.

\section{Code Availability}\label{sec6}

Code used to reproduce the analyses presented here can be found at \url{https://github.com/npbyers0401/citation-function-analysis}.

\section{Acknowledgements}\label{sec7}

The work conducted by the U.S. Department of Energy Joint Genome Institute (\url{https://ror.org/04xm1d337}), a DOE Office of Science User Facility, is supported by the Office of Science of the U.S. Department of Energy operated under Contract No. DE-AC02-05CH11231. This research used the CBorg AI platform and the Lawrencium computational cluster resource provided by the IT Division at the Lawrence Berkeley National Laboratory (Supported by the Director, Office of Science, Office of Basic Energy Sciences, of the U.S. Department of Energy under Contract No. DE-AC02-05CH11231).

\section{Statements and Declarations}\label{sec8}

\subsection{Competing Interests}\label{sec8a}

The authors have no competing interests to declare that are relevant to the content of this article.

\nocite{*} 


\begin{thebibliography}{88}
\providecommand{\natexlab}[1]{#1}
\providecommand{\url}[1]{{#1}}
\providecommand{\urlprefix}{URL }
\providecommand{\doi}[1]{\url{https://doi.org/#1}}
\providecommand{\eprint}[2][]{\url{#2}}
 \bibcommenthead

\bibitem[{Aati et~al.(2022)Aati, Perveen, Aati, Orfali, Alqahtani, Al-Taweel, Wanner, and Aati}]{aati_headspace_2022}
Aati HY, Perveen S, Aati S, et~al (2022) Headspace solid-phase microextraction method for extracting volatile constituents from the different parts of saudi anethum graveolens l. and their antimicrobial activity. Heliyon 8(3):e09051. \doi{10.1016/j.heliyon.2022.e09051}

\bibitem[{Agrawal et~al.(2025)Agrawal, Tan, Soylu, Ziems, Khare, Opsahl-Ong, Singhvi, Shandilya, Ryan, Jiang, Potts, Sen, Dimakis, Stoica, Klein, Zaharia, and Khattab}]{agrawal2025gepareflectivepromptevolution}
Agrawal LA, Tan S, Soylu D, et~al (2025) {GEPA}: Reflective prompt evolution can outperform reinforcement learning. \doi{10.48550/arXiv.2507.19457}, [Preprint]

\bibitem[{Alex and Antunes(2019)}]{alex_comparative_2019}
Alex A, Antunes A (2019) Comparative genomics reveals metabolic specificity of endozoicomonas isolated from a marine sponge and the genomic repertoire for host-bacteria symbioses. Microorganisms 7(12):635. \doi{10.3390/microorganisms7120635}

\bibitem[{Anthropic(2025)}]{anthropic_claude_2025}
Anthropic (2025) Claude \uppercase{O}pus 4.1. \urlprefix\url{https://www.anthropic.com/news/claude-opus-4-1}, retrieved October 20, 2025

\bibitem[{Bansal et~al.(2024)Bansal, Banda, Glatt-Deeley, Stoddard, Linsley, Arora, Deleschaux, Ahern, Kondaveeti, Massey, Nicouleau, Wang, Sabariego-Navarro, Dierssen, Finkbeiner, and Pinter}]{bansal_dynamic_2024}
Bansal P, Banda EC, Glatt-Deeley HR, et~al (2024) A dynamic in vitro model of down syndrome neurogenesis with trisomy 21 gene dosage correction. Science Advances 10(23):eadj0385. \doi{10.1126/sciadv.adj0385}

\bibitem[{Branger et~al.(2016)Branger, Hauer, Michelet, Karoui, Cochard, De~Cruz, Henault, Boschiroli, and Biet}]{branger_draft_2016}
Branger M, Hauer A, Michelet L, et~al (2016) Draft genome sequence of mycobacterium bovis strain d-10-02315 isolated from wild boar. Genome Announcements 4(6):e01268--16. \doi{10.1128/genomea.01268-16}

\bibitem[{Budi and Yaniasih(2022)}]{budi_understanding_2022}
Budi I, Yaniasih Y (2022) Understanding the meanings of citations using sentiment, role, and citation function classifications. Scientometrics 128(1):735--759. \doi{10.1007/s11192-022-04567-4}

\bibitem[{Burrichter et~al.(2021)Burrichter, Dörr, Bergmann, Haiß, Keller, Fournier, Franchini, Isono, and Schleheck}]{burrichter_bacterial_2021}
Burrichter AG, Dörr S, Bergmann P, et~al (2021) Bacterial microcompartments for isethionate desulfonation in the taurine-degrading human-gut bacterium bilophila wadsworthia. {BMC} Microbiology 21(1):340. \doi{10.1186/s12866-021-02386-w}

\bibitem[{Byers et~al.(2024)Byers, Parker, Beecroft, Reddy, Salamon, Garrity, and Fagnan}]{byers_identifying_2024}
Byers N, Parker C, Beecroft C, et~al (2024) Identifying genomic data use with the {D}ata {C}itation {E}xplorer. Scientific Data 11(1):1200. \doi{10.1038/s41597-024-04049-7}

\bibitem[{Chatrinan et~al.(2025)Chatrinan, Noraset, and Tuarob}]{chatrinan_overcoming_2025}
Chatrinan K, Noraset T, Tuarob S (2025) Overcoming data scarcity: Guiding citation function classification with prompt-based few-shot learning. {IEEE} Access 13:119188--119196. \doi{10.1109/access.2025.3586729}

\bibitem[{Chen et~al.(2024)Chen, Chen, Zhou, Shang, Tang, Xu, Duan, Jin, Xu, Yan, and Chen}]{chen_two_2024}
Chen HX, Chen FJ, Zhou QJ, et~al (2024) Two colistin resistance-producing aeromonas strains, isolated from coastal waters in zhejiang, china: characteristics, multi-drug resistance and pathogenicity. Frontiers in Microbiology 15:1401802. \doi{10.3389/fmicb.2024.1401802}

\bibitem[{Chen et~al.(2022)Chen, Chu, Palaniappan, Ratner, Huang, Huntemann, Hajek, Ritter, Webb, Wu, Varghese, Reddy, Mukherjee, Ovchinnikova, Nolan, Seshadri, Roux, Visel, Woyke, Eloe-Fadrosh, Kyrpides, and Ivanova}]{chen_imgm_2022}
Chen IMA, Chu K, Palaniappan K, et~al (2022) The {IMG}/{M} data management and analysis system v.7: content updates and new features. Nucleic Acids Research 51:d723--d732. \doi{10.1093/nar/gkac976}

\bibitem[{Chorlton(2024)}]{chorlton_ten_2024}
Chorlton SD (2024) Ten common issues with reference sequence databases and how to mitigate them. Frontiers in Bioinformatics 4:1278228. \doi{10.3389/fbinf.2024.1278228}

\bibitem[{Chotoo et~al.(2013)Chotoo, Silverman, Devor, and Luke}]{chotoo_small_2013}
Chotoo CK, Silverman GA, Devor DC, et~al (2013) A small conductance calcium-activated k+ channel in c. elegans, {KCNL}-2, plays a role in the regulation of the rate of egg-laying. {PLoS} {ONE} 8(9):e75869. \doi{10.1371/journal.pone.0075869}

\bibitem[{Clabaut et~al.(2019)Clabaut, Boukerb, Racine, Pichon, Kremser, Queiroz, Karsybayeva, Redziniak, Chevalier, and Feuilloley}]{clabaut_draft_2019}
Clabaut M, Boukerb AM, Racine PJ, et~al (2019) Draft genome sequence of lactobacillus crispatus strain v4, isolated from a vaginal swab from a young healthy nonmenopausal woman. Microbiology Resource Announcements 8(38). \doi{10.1128/mra.00856-19}

\bibitem[{Clark et~al.(2018)Clark, Cowhey, Etzioni, Khot, Sabharwal, Schoenick, and Tafjord}]{clark_think_2018}
Clark P, Cowhey I, Etzioni O, et~al (2018) Think you have solved question answering? try {ARC}, the {AI}2 reasoning challenge. \doi{10.48550/arxiv.1803.05457}, [Preprint]

\bibitem[{Cohan et~al.(2019)Cohan, Ammar, van Zuylen, and Cady}]{cohan_structural_2019}
Cohan A, Ammar W, van Zuylen M, et~al (2019) Structural scaffolds for citation intent classification in scientific publications. In: Proceedings of the 2019 Conference of the North, pp 3586--3596, \doi{10.18653/v1/n19-1361}

\bibitem[{Cousijn et~al.(2019)Cousijn, Feeney, Lowenberg, Presani, and Simons}]{cousijn_bringing_2019}
Cousijn H, Feeney P, Lowenberg D, et~al (2019) Bringing citations and usage metrics together to make data count. Data Science Journal 18:9. \doi{10.5334/dsj-2019-009}

\bibitem[{Dimensions(2025)}]{ds_dimensions_2025}
Dimensions (2025) Dimensions: The most comprehensive view of the research landscape. \urlprefix\url{https://www.dimensions.ai/products/}, retrieved October 20, 2025

\bibitem[{Dong and Schäfer(2011)}]{dong_ensemble-style_2011}
Dong C, Schäfer U (2011) Ensemble-style self-training on citation classification. In: Proceedings of 5th International Joint Conference on Natural Language Processing, pp 623--631, \urlprefix\url{https://aclanthology.org/I11-1070/}

\bibitem[{Etges et~al.(2016)Etges, de~Oliveira, Rajpurohit, and Gibbs}]{etges_effects_2016}
Etges WJ, de~Oliveira CC, Rajpurohit S, et~al (2016) Effects of temperature on transcriptome and cuticular hydrocarbon expression in ecologically differentiated populations of desert drosophila. Ecology and Evolution 7(2):619--637. \doi{10.1002/ece3.2653}

\bibitem[{Farmer et~al.(2023)Farmer, Rajasabhai, Tarpeh, Tyo, and Wells}]{farmer_meta-omic_2023}
Farmer M, Rajasabhai R, Tarpeh W, et~al (2023) Meta-omic profiling reveals ubiquity of genes encoding for the nitrogen-rich biopolymer cyanophycin in activated sludge microbiomes. Frontiers in Microbiology 14:1287491. \doi{10.3389/fmicb.2023.1287491}

\bibitem[{Frisch et~al.(2018)Frisch, Castillo-Ramírez, Petit, Farley, Ray, Albrecht, Limbago, Hernandez, See, Satola, and Read}]{frisch_invasive_2018}
Frisch MB, Castillo-Ramírez S, Petit RA, et~al (2018) Invasive methicillin-resistant staphylococcus aureus {USA}500 strains from the u.s. emerging infections program constitute three geographically distinct lineages. {mSphere} 3(3):10.1128/msphere.00571--17. \doi{10.1128/msphere.00571-17}

\bibitem[{Goodstein et~al.(2011)Goodstein, Shu, Howson, Neupane, Hayes, Fazo, Mitros, Dirks, Hellsten, Putnam, and Rokhsar}]{goodstein_phytozome_2011}
Goodstein DM, Shu S, Howson R, et~al (2011) Phytozome: a comparative platform for green plant genomics. Nucleic Acids Research 40:d1178--d1186. \doi{10.1093/nar/gkr944}

\bibitem[{Gowda et~al.(2024)Gowda, Raj, and Madasamy}]{c_citation_2024}
Gowda RHC, Raj KS, Madasamy AK (2024) Citation intent classification using transformers. In: 2024 {IEEE} Students Conference on Engineering and Systems ({SCES}), pp 1--6, \doi{10.1109/sces61914.2024.10652428}

\bibitem[{Grattafiori et~al.(2024)Grattafiori, Dubey, Jauhri, Pandey, Kadian, Al-Dahle, Letman, Mathur, Schelten, Vaughan, Yang, Fan, Goyal, Hartshorn, Yang, Mitra, Sravankumar, Korenev, Hinsvark, Rao, Zhang, Rodriguez, Gregerson, Spataru, Roziere, Biron, Tang, Chern, Caucheteux, Nayak, Bi, Marra, McConnell, Keller, Touret, Wu, Wong, Ferrer, Nikolaidis, Allonsius, Song, Pintz, Livshits, Wyatt, Esiobu, Choudhary, Mahajan, Garcia-Olano, Perino, Hupkes, Lakomkin, AlBadawy, Lobanova, Dinan, Smith, Radenovic, Guzmán, Zhang, Synnaeve, Lee, Anderson, Thattai, Nail, Mialon, Pang, Cucurell, Nguyen, Korevaar, Xu, Touvron, Zarov, Ibarra, Kloumann, Misra, Evtimov, Zhang, Copet, Lee, Geffert, Vranes, Park, Mahadeokar, Shah, van~der Linde, Billock, Hong, Lee, Fu, Chi, Huang, Liu, Wang, Yu, Bitton, Spisak, Park, Rocca, Johnstun, Saxe, Jia, Alwala, Prasad, Upasani, Plawiak, Li, Heafield, Stone, El-Arini, Iyer, Malik, Chiu, Bhalla, Lakhotia, Rantala-Yeary, van~der Maaten, Chen, Tan, Jenkins, Martin, Madaan, Malo, Blecher,
  Landzaat, de~Oliveira, Muzzi, Pasupuleti, Singh, Paluri, Kardas, Tsimpoukelli, Oldham, Rita, Pavlova, Kambadur, Lewis, Si, Singh, Hassan, Goyal, Torabi, Bashlykov, Bogoychev, Chatterji, Zhang, Duchenne, Çelebi, Alrassy, Zhang, Li, Vasic, Weng, Bhargava, Dubal, Krishnan, Koura, Xu, He, Dong, Srinivasan, Ganapathy, Calderer, Cabral, Stojnic, Raileanu, Maheswari, Girdhar, Patel, Sauvestre, Polidoro, Sumbaly, Taylor, Silva, Hou, Wang, Hosseini, Chennabasappa, Singh, Bell, Kim, Edunov, Nie, Narang, Raparthy, Shen, Wan, Bhosale, Zhang, Vandenhende, Batra, Whitman, Sootla, Collot, Gururangan, Borodinsky, Herman, Fowler, Sheasha, Georgiou, Scialom, Speckbacher, Mihaylov, Xiao, Karn, Goswami, Gupta, Ramanathan, Kerkez, Gonguet, Do, Vogeti, Albiero, Petrovic, Chu, Xiong, Fu, Meers, Martinet, Wang, Wang, Tan, Xia, Xie, Jia, Wang, Goldschlag, Gaur, Babaei, Wen, Song, Zhang, Li, Mao, Coudert, Yan, Chen, Papakipos, Singh, Srivastava, Jain, Kelsey, Shajnfeld, Gangidi, Victoria, Goldstand, Menon, Sharma, Boesenberg,
  Baevski, Feinstein, Kallet, Sangani, Teo, Yunus, Lupu, Alvarado, Caples, Gu, Ho, Poulton, Ryan, Ramchandani, Dong, Franco, Goyal, Saraf, Chowdhury, Gabriel, Bharambe, Eisenman, Yazdan, James, Maurer, Leonhardi, Huang, Loyd, Paola, Paranjape, Liu, Wu, Ni, Hancock, Wasti, Spence, Stojkovic, Gamido, Montalvo, Parker, Burton, Mejia, Liu, Wang, Kim, Zhou, Hu, Chu, Cai, Tindal, Feichtenhofer, Gao, Civin, Beaty, Kreymer, Li, Adkins, Xu, Testuggine, David, Parikh, Liskovich, Foss, Wang, Le, Holland, Dowling, Jamil, Montgomery, Presani, Hahn, Wood, Le, Brinkman, Arcaute, Dunbar, Smothers, Sun, Kreuk, Tian, Kokkinos, Ozgenel, Caggioni, Kanayet, Seide, Florez, Schwarz, Badeer, Swee, Halpern, Herman, Sizov, Guangyi, Zhang, Lakshminarayanan, Inan, Shojanazeri, Zou, Wang, Zha, Habeeb, Rudolph, Suk, Aspegren, Goldman, Zhan, Damlaj, Molybog, Tufanov, Leontiadis, Veliche, Gat, Weissman, Geboski, Kohli, Lam, Asher, Gaya, Marcus, Tang, Chan, Zhen, Reizenstein, Teboul, Zhong, Jin, Yang, Cummings, Carvill, Shepard, McPhie,
  Torres, Ginsburg, Wang, Wu, U, Saxena, Khandelwal, Zand, Matosich, Veeraraghavan, Michelena, Li, Jagadeesh, Huang, Chawla, Huang, Chen, Garg, A, Silva, Bell, Zhang, Guo, Yu, Moshkovich, Wehrstedt, Khabsa, Avalani, Bhatt, Mankus, Hasson, Lennie, Reso, Groshev, Naumov, Lathi, Keneally, Liu, Seltzer, Valko, Restrepo, Patel, Vyatskov, Samvelyan, Clark, Macey, Wang, Hermoso, Metanat, Rastegari, Bansal, Santhanam, Parks, White, Bawa, Singhal, Egebo, Usunier, Mehta, Laptev, Dong, Cheng, Chernoguz, Hart, Salpekar, Kalinli, Kent, Parekh, Saab, Balaji, Rittner, Bontrager, Roux, Dollar, Zvyagina, Ratanchandani, Yuvraj, Liang, Alao, Rodriguez, Ayub, Murthy, Nayani, Mitra, Parthasarathy, Li, Hogan, Battey, Wang, Howes, Rinott, Mehta, Siby, Bondu, Datta, Chugh, Hunt, Dhillon, Sidorov, Pan, Mahajan, Verma, Yamamoto, Ramaswamy, Lindsay, Lindsay, Feng, Lin, Zha, Patil, Shankar, Zhang, Zhang, Wang, Agarwal, Sajuyigbe, Chintala, Max, Chen, Kehoe, Satterfield, Govindaprasad, Gupta, Deng, Cho, Virk, Subramanian, Choudhury,
  Goldman, Remez, Glaser, Best, Koehler, Robinson, Li, Zhang, Matthews, Chou, Shaked, Vontimitta, Ajayi, Montanez, Mohan, Kumar, Mangla, Ionescu, Poenaru, Mihailescu, Ivanov, Li, Wang, Jiang, Bouaziz, Constable, Tang, Wu, Wang, Wu, Gao, Kleinman, Chen, Hu, Jia, Qi, Li, Zhang, Zhang, Adi, Nam, Yu, Wang, Zhao, Hao, Qian, Li, He, Rait, DeVito, Rosnbrick, Wen, Yang, Zhao, and Ma}]{grattafiori_llama_2024}
Grattafiori A, Dubey A, Jauhri A, et~al (2024) The \uppercase{L}lama 3 herd of models. \doi{10.48550/arXiv.2407.21783}, [Preprint]

\bibitem[{Grigoriev et~al.(2013)Grigoriev, Nikitin, Haridas, Kuo, Ohm, Otillar, Riley, Salamov, Zhao, Korzeniewski, Smirnova, Nordberg, Dubchak, and Shabalov}]{grigoriev_mycocosm_2013}
Grigoriev IV, Nikitin R, Haridas S, et~al (2013) {MycoCosm} portal: gearing up for 1000 fungal genomes. Nucleic Acids Research 42:d699--d704. \doi{10.1093/nar/gkt1183}

\bibitem[{Grigoriev et~al.(2020)Grigoriev, Hayes, Calhoun, Kamel, Wang, Ahrendt, Dusheyko, Nikitin, Mondo, Salamov, Shabalov, and Kuo}]{grigoriev_phycocosm_2020}
Grigoriev IV, Hayes RD, Calhoun S, et~al (2020) {PhycoCosm}, a comparative algal genomics resource. Nucleic Acids Research 49:d1004--d1011. \doi{10.1093/nar/gkaa898}

\bibitem[{Guha et~al.(2023)Guha, Nyarko, Ho, Ré, Chilton, Narayana, Chohlas-Wood, Peters, Waldon, Rockmore, Zambrano, Talisman, Hoque, Surani, Fagan, Sarfaty, Dickinson, Porat, Hegland, Wu, Nudell, Niklaus, Nay, Choi, Tobia, Hagan, Ma, Livermore, Rasumov-Rahe, Holzenberger, Kolt, Henderson, Rehaag, Goel, Gao, Williams, Gandhi, Zur, Iyer, and Li}]{guha_legalbench_2023}
Guha N, Nyarko J, Ho DE, et~al (2023) {LegalBench}: A collaboratively built benchmark for measuring legal reasoning in large language models. \doi{10.48550/arxiv.2308.11462}, [Preprint]

\bibitem[{Göker et~al.(2012)Göker, Saunders, Lapidus, Nolan, Lucas, Hammon, Deshpande, Cheng, Han, Tapia, Goodwin, Pitluck, Liolios, Mavromatis, Pagani, Ivanova, Mikhailova, Pati, Chen, Palaniappan, Land, Chang, Jeffries, Brambilla, Rohde, Spring, Detter, Woyke, Bristow, Eisen, Markowitz, Hugenholtz, Kyrpides, and Klenk}]{goker_genome_2012}
Göker M, Saunders E, Lapidus A, et~al (2012) Genome sequence of the moderately thermophilic, amino-acid-degrading and sulfur-reducing bacterium thermovirga lienii type strain (cas60314t). Standards in Genomic Sciences 6(2):230--239. \doi{10.4056/sigs.2726028}

\bibitem[{Hazards et~al.(2022)Hazards, Koutsoumanis, Allende, Alvarez‐Ordóñez, Bolton, Bover‐Cid, Chemaly, Davies, De~Cesare, Hilbert, Lindqvist, Nauta, Peixe, Ru, Simmons, Skandamis, Suffredini, Cocconcelli, Escámez, Maradona, Querol, Sijtsma, Suarez, Sundh, Vlak, Barizzone, Hempen, Correia, and Herman}]{hazards_update_2022}
Hazards EPoB, Koutsoumanis K, Allende A, et~al (2022) Update of the list of {QPS}‐recommended microbiological agents intentionally added to food or feed as notified to {EFSA} 16: suitability of taxonomic units notified to {EFSA} until march 2022. {EFSA} Journal 20(7):e07408. \doi{10.2903/j.efsa.2022.7408}

\bibitem[{Hendricks et~al.(2020)Hendricks, Tkaczyk, Lin, and Feeney}]{hendricks_crossref_2020}
Hendricks G, Tkaczyk D, Lin J, et~al (2020) Crossref: The sustainable source of community-owned scholarly metadata. Quantitative Science Studies 1(1):414--427. \doi{10.1162/qss_a_00022}

\bibitem[{Hermanns et~al.(2019)Hermanns, Marklewitz, Zirkel, Overheul, Page, Loaiza, Drosten, van Rij, and Junglen}]{hermanns_agua_2019}
Hermanns K, Marklewitz M, Zirkel F, et~al (2019) Agua salud alphavirus defines a novel lineage of insect-specific alphaviruses discovered in the new world. The Journal of General Virology 101(1):96--104. \doi{10.1099/jgv.0.001344}

\bibitem[{{Hernández}-{Alvarez} et~al.(2017){Hernández}-{Alvarez}, {Soriano}, and {Martínez}-{Barco}}]{hernandez-alvarez_citation_2017}
{Hernández}-{Alvarez} M, {Soriano} JMG, {Martínez}-{Barco} P (2017) Citation function, polarity and influence classification. Natural Language Engineering 23(4):561--588. \doi{10.1017/s1351324916000346}

\bibitem[{Hong et~al.(2025)Hong, Troynikov, and Huber}]{hong_context_2025}
Hong K, Troynikov A, Huber J (2025) Context rot: How increasing input tokens impacts llm performance. \urlprefix\url{https://research.trychroma.com/context-rot}, retrieved October 20, 2025

\bibitem[{Howe et~al.(2024)Howe, Zaugg, and Mason}]{howe_novel_2024}
Howe KL, Zaugg J, Mason OU (2024) Novel, active, and uncultured hydrocarbon-degrading microbes in the ocean. Applied and Environmental Microbiology 90(9):e01224--24. \doi{10.1128/aem.01224-24}

\bibitem[{Ichida and Long(2016)}]{ichida_ldss-p_2016}
Ichida H, Long SR (2016) {LDSS}-p: an advanced algorithm to extract functional short motifs associated with coordinated gene expression. Nucleic Acids Research 44(11):5045--5053. \doi{10.1093/nar/gkw435}

\bibitem[{ICPSR(2025)}]{icpsr_bib_2025}
ICPSR (2025) Bibliography of data-related literature: Data-linked publication citation files, full download. \doi{10.3886/E218241V2}, [Dataset] Ann Arbor, MI: Inter-university Consortium for Political and Social Research

\bibitem[{Jagadeesan et~al.(2019)Jagadeesan, Meenakshisundaram, Boopathy, Mookandi, and Balaiah}]{jagadeesan_combinatorial_2019}
Jagadeesan Y, Meenakshisundaram S, Boopathy LRA, et~al (2019) Combinatorial approach for screening and assessment of multiple therapeutic enzymes from marine isolate pseudomonas aeruginosa {AR}01. {RSC} Advances 9(30):16989--17001. \doi{10.1039/c9ra02555c}

\bibitem[{JGI(2025)}]{jgi_portal_2025}
JGI (2025) \uppercase{JGI} \uppercase{D}ata \uppercase{P}ortal. \urlprefix\url{https://data.jgi.doe.gov}, retrieved October 20, 2025

\bibitem[{{Jha} et~al.(2016){Jha}, {Jbara}, {Qazvinian}, and {Radev}}]{jha_nlp-driven_2016}
{Jha} R, {Jbara} AA, {Qazvinian} V, et~al (2016) {NLP}-driven citation analysis for scientometrics. Natural Language Engineering 23(1):93--130. \doi{10.1017/s1351324915000443}

\bibitem[{Ji et~al.(2021)Ji, Byun, Chen, Ahn, Jung, Han, Lim, Won, Moon, Hur, and Seo}]{ji_radiation-inactivated_2021}
Ji HJ, Byun EB, Chen F, et~al (2021) Radiation-inactivated s. gallinarum vaccine provides a high protective immune response by activating both humoral and cellular immunity. Frontiers in Immunology 12:717556. \doi{10.3389/fimmu.2021.717556}

\bibitem[{Jiang and Chen(2023)}]{jiang_contextualised_2023}
Jiang X, Chen J (2023) Contextualised segment-wise citation function classification. Scientometrics 128(9):5117--5158. \doi{10.1007/s11192-023-04778-3}

\bibitem[{Jurgens et~al.(2018)Jurgens, Kumar, Hoover, {McFarland}, and Jurafsky}]{jurgens_measuring_2018}
Jurgens D, Kumar S, Hoover R, et~al (2018) Measuring the evolution of a scientific field through citation frames. Transactions of the Association for Computational Linguistics 6:391--406. \doi{10.1162/tacl_a_00028}

\bibitem[{Lam et~al.(2024)Lam, Kajderowicz, Keys, Roessler, Frenkel, Kirkland, Bisht, El-Brolosy, Jaenisch, Bell, Weissman, Griffith, and Hrvatin}]{lam_multi-species_2024}
Lam B, Kajderowicz KM, Keys HR, et~al (2024) Multi-species genome-wide {CRISPR} screens identify conserved suppressors of cold-induced cell death. {bioRxiv} p 2024.07.25.605098. \doi{10.1101/2024.07.25.605098}

\bibitem[{Leavitt et~al.(2024)Leavitt, Woodbury, Gilcrease, Bridges, Teschke, and Casjens}]{leavitt_bacteriophage_2024}
Leavitt JC, Woodbury BM, Gilcrease EB, et~al (2024) Bacteriophage p22 {SieA}-mediated superinfection exclusion. {mBio} 15(2):e02169--23. \doi{10.1128/mbio.02169-23}

\bibitem[{Lee et~al.(2020)Lee, Badaya, Singh, and Lim}]{lee_complete_2020}
Lee K, Badaya SK, Singh R, et~al (2020) Complete genome sequence of gordonia sp. strain {JH}63, isolated from human skin. Microbiology Resource Announcements 9(11):10.1128/mra.00059--20. \doi{10.1128/mra.00059-20}

\bibitem[{Lewis et~al.(2021)Lewis, Perez, Piktus, Petroni, Karpukhin, Goyal, Küttler, Lewis, tau Yih, Rocktäschel, Riedel, and Kiela}]{lewis2021retrievalaugment}
Lewis P, Perez E, Piktus A, et~al (2021) Retrieval-augmented generation for knowledge-intensive nlp tasks. \doi{10.48550/arXiv.2005.11401}, [Preprint]

\bibitem[{Li and Du(2012)}]{li_gene_2012}
Li XQ, Du D (2012) Gene direction in living organisms. Scientific Reports 2(1):982. \doi{10.1038/srep00982}

\bibitem[{Liu et~al.(2023)Liu, Xiong, Jiang, Ma, Lu, Huang, and Cheng}]{liu_low-resource_2023}
Liu J, Xiong Z, Jiang Y, et~al (2023) Low-resource multi-granularity academic function recognition based on multiple prompt knowledge. \doi{10.48550/arxiv.2305.03287}, [Preprint]

\bibitem[{Lo et~al.(2020)Lo, Wang, Neumann, Kinney, and Weld}]{lo-etal-2020-s2orc}
Lo K, Wang LL, Neumann M, et~al (2020) {S}2{ORC}: The semantic scholar open research corpus. In: Proceedings of the 58th Annual Meeting of the Association for Computational Linguistics, pp 4969--4983, \doi{"10.18653/v1/2020.acl-main.447}

\bibitem[{Machado and Gram(2017)}]{machado_comparative_2017}
Machado H, Gram L (2017) Comparative genomics reveals high genomic diversity in the genus photobacterium. Frontiers in Microbiology 8:1204. \doi{10.3389/fmicb.2017.01204}

\bibitem[{Matsutani et~al.(2013)Matsutani, Kawajiri, Yakushi, Adachi, and Matsushita}]{matsutani_draft_2013}
Matsutani M, Kawajiri E, Yakushi T, et~al (2013) Draft genome sequence of dihydroxyacetone-producing gluconobacter thailandicus strain {NBRC} 3255. Genome Announcements 1(2):e00118--13. \doi{10.1128/genomea.00118-13}

\bibitem[{Moon et~al.(2017)Moon, Kong, Hong, Lee, and Quan}]{moon_identification_2017}
Moon EK, Kong HH, Hong Y, et~al (2017) Identification and characterization of protein arginine methyltransferase 1 in \textit{Acanthamoeba castellanii}. The Korean Journal of Parasitology 55(2):109--114. \doi{10.3347/kjp.2017.55.2.109}

\bibitem[{Moreno et~al.(2010)Moreno, Marinotti, Krzywinski, Tadei, James, Achee, and Conn}]{moreno_complete_2010}
Moreno M, Marinotti O, Krzywinski J, et~al (2010) Complete {mtDNA} genomes of anopheles darlingi and an approach to anopheline divergence time. Malaria Journal 9(1):127. \doi{10.1186/1475-2875-9-127}

\bibitem[{Mukherjee et~al.(2016)Mukherjee, Stamatis, Bertsch, Ovchinnikova, Verezemska, Isbandi, Thomas, Ali, Sharma, Kyrpides, and Reddy}]{mukherjee_genomes_2016}
Mukherjee S, Stamatis D, Bertsch J, et~al (2016) Genomes {OnLine} database ({GOLD}) v.6: data updates and feature enhancements. Nucleic Acids Research 45:d446--d456. \doi{10.1093/nar/gkw992}

\bibitem[{Nambanoor~Kunnath(2024)}]{nambanoor_language_2024}
Nambanoor~Kunnath S (2024) Language models for citation classification. PhD thesis, The Open University, \doi{10.21954/ou.ro.00095746}

\bibitem[{Nedashkovskaya et~al.(2024)Nedashkovskaya, Balabanova, Otstavnykh, Zhukova, Detkova, Seitkalieva, Bystritskaya, Noskova, Tekutyeva, and Isaeva}]{nedashkovskaya_-depth_2024}
Nedashkovskaya O, Balabanova L, Otstavnykh N, et~al (2024) In-depth genome characterization and pan-genome analysis of strain {KMM} 296, a producer of highly active alkaline phosphatase; proposal for the reclassification of cobetia litoralis and cobetia pacifica as the later heterotypic synonyms of cobetia amphilecti and cobetia marina, and emended description of the species cobetia amphilecti and cobetia marina. Biomolecules 14(2):196. \doi{10.3390/biom14020196}

\bibitem[{Neumann and Brase(2014)}]{neumann_datacite_2014}
Neumann J, Brase J (2014) {DataCite} and {DOI} names for research data. Journal of Computer-Aided Molecular Design 28(10):1035--1041. \doi{10.1007/s10822-014-9776-5}

\bibitem[{Opsahl-Ong et~al.(2024)Opsahl-Ong, Ryan, Purtell, Broman, Potts, Zaharia, and Khattab}]{opsahlong2024optimizing}
Opsahl-Ong K, Ryan MJ, Purtell J, et~al (2024) Optimizing instructions and demonstrations for multi-stage language model programs. \doi{10.48550/arXiv.2406.11695}, [Preprint]

\bibitem[{Otun and Ntushelo(2023)}]{otun_data_2023}
Otun SO, Ntushelo K (2023) A data catalogue of alcohol dehydrogenases in the genome of a tissue macerating pectobacterium brasiliense isolated from potato in south africa. Data in Brief 52:109918. \doi{10.1016/j.dib.2023.109918}

\bibitem[{Phan et~al.(2025)Phan, Gatti, Han, Li, Hu, Zhang, Zhang, Shaaban, Ling, Shi, Choi, Agrawal, Chopra, Khoja, Kim, Ren, Hausenloy, Zhang, Mazeika, Dodonov, Nguyen, Lee, Anderson, Doroshenko, Stokes, Mahmood, Pokutnyi, Iskra, Wang, Levin, Kazakov, Feng, Feng, Zhao, Yu, Gangal, Zou, Wang, Popov, Gerbicz, Galgon, Schmitt, Yeadon, Lee, Sauers, Sanchez, Giska, Roth, Riis, Utpala, Burns, Goshu, Naiya, Agu, Giboney, Cheatom, Fournier-Facio, Crowson, Finke, Cheng, Zampese, Hoerr, Nandor, Park, Gehrunger, Cai, McCarty, Garretson, Taylor, Sileo, Ren, Qazi, Li, Nam, Wydallis, Arkhipov, Shi, Bacho, Willcocks, Cao, Motwani, de~Oliveira~Santos, Veith, Vendrow, Cojoc, Zenitani, Robinson, Tang, Li, Vendrow, Fraga, Kuchkin, Maksimov, Marion, Efremov, Lynch, Liang, Mikov, Gritsevskiy, Guillod, Demir, Martinez, Pageler, Zhou, Soori, Press, Tang, Rissone, Green, Brüssel, Twayana, Dieuleveut, Imperial, Prabhu, Yang, Crispino, Rao, Zvonkine, Loiseau, Kalinin, Lukas, Manolescu, Stambaugh, Mishra, Hogg, Bosio, Coppola,
  Salazar, Jin, Sayous, Ivanov, Schwaller, Senthilkuma, Bran, Algaba, den Houte, Sypt, Verbeken, Noever, Kopylov, Myklebust, Li, Schut, Zheltonozhskii, Yuan, Lim, Stanley, Yang, Maar, Wykowski, Oller, Sahu, Ardito, Hu, Kamdoum, Jin, Vilchis, Zu, Lackner, Koppel, Sun, Antonenko, Chern, Zhao, Arsene, Cavanagh, Li, Shen, Crisostomi, Zhang, Dehghan, Ivanov, Perrella, Kaparov, Zang, Sucholutsky, Kharlamova, Orel, Poritski, Ben-David, Berger, Whitfill, Foster, Munro, Ho, Sivarajan, Hava, Kuchkin, Holmes, Rodriguez-Romero, Sommerhage, Zhang, Moat, Schneider, Kazibwe, Clarke, Kim, Dias, Fish, Elser, Kreiman, Vilchis, Klose, Anantheswaran, Zweiger, Rawal, Li, Nguyen, Daans, Heidinger, Radionov, Rozhoň, Ginis, Stump, Cohen, Poświata, Tkadlec, Goldfarb, Wang, Padlewski, Barzowski, Montgomery, Stendall, Tucker-Foltz, Stade, Rogers, Goertzen, Grabb, Shukla, Givré, Ambay, Sen, Aziz, Inlow, He, Zhang, Kaddar, Ängquist, Chen, Wang, Ramakrishnan, Thornley, Terpin, Schoelkopf, Zheng, Carmi, Brown, Zhu, Bartolo, Wheeler,
  Stehberger, Bradshaw, Heimonen, Sridhar, Akov, Sandlin, Makarychev, Tam, Hoang, Cunningham, Goryachev, Patramanis, Krause, Redenti, Aldous, Lai, Coleman, Xu, Lee, Magoulas, Zhao, Tang, Cohen, Paradise, Kirchner, Ovchynnikov, Matos, Shenoy, Wang, Nie, Sztyber-Betley, Faraboschi, Riblet, Crozier, Halasyamani, Verma, Joshi, Meril, Ma, Andréoletti, Singhal, Platnick, Nevirkovets, Basler, Ivanov, Khoury, Gustafsson, Piccardo, Mostaghimi, Chen, Singh, Khánh, Rosu, Szlyk, Brown, Narayan, Menezes, Roberts, Alley, Sun, Patel, Lamparth, Reuel, Xin, Xu, Loader, Martin, Wang, Achilleos, Preu, Korbak, Bosio, Kazemi, Chen, Bálint, Lo, Wang, Nunes, Milbauer, Bari, Wang, Ansarinejad, Sun, Durand, Elgnainy, Douville, Tordera, Balabanian, Wolff, Kvistad, Milliron, Sakor, Eron, O., Shah, Zhou, Kamalov, Abdoli, Santens, Barkan, Tee, Zhang, Tomasiello, Luca, Looi, Le, Kolt, Pan, Rodman, Drori, Fossum, Muennighoff, Jagota, Pradeep, Fan, Eicher, Chen, Thaman, Merrill, Firsching, Harris, Ciobâcă, Gross, Pandey, Gusev, Jones,
  Agnihotri, Zhelnov, Mofayezi, Piperski, Zhang, Dobarskyi, Leventov, Soroko, Duersch, Taamazyan, Ho, Ma, Held, Xian, Zebaze, Mohamed, Leser, Yuan, Yacar, Lengler, Olszewska, Fratta, Oliveira, Jackson, Zou, Chidambaram, Manik, Haffenden, Stander, Dasouqi, Shen, Golshani, Stap, Kretov, Uzhou, Zhidkovskaya, Winter, Rodriguez, Lauff, Wehr, Tang, Hossain, Phillips, Samuele, Ekström, Hammon, Patel, Farhidi, Medley, Mohammadzadeh, Peñaflor, Kassahun, Friedrich, Perez, Pyda, Sakal, Dhamane, Mirabadi, Hallman, Okutsu, Battaglia, Maghsoudimehrabani, Amit, Hulbert, Pereira, Weber, Handoko, Peristyy, Malina, Mehkary, Aly, Reidegeld, Dick, Friday, Singh, Shapourian, Kim, Costa, Gurdogan, Kumar, Ceconello, Zhuang, Park, Carroll, Tawfeek, Steinerberger, Aggarwal, Kirchhof, Dai, Kim, Ferret, Shah, Wang, Yan, Burdzy, Zhang, Franca, Pham, Loh, Robinson, Jackson, Giordano, Petersen, Cosma, Colino, White, Votava, Vinnikov, Delaney, Spelda, Stritecky, Shahid, Mourrat, Vetoshkin, Sponselee, Bacho, Yong, de~la Rosa, Cho, Li,
  Malod, Weller, Albani, Lang, Laurendeau, Kazakov, Adesanya, Portier, Hollom, Souza, Zhou, Degorre, Yalın, Obikoya, Rai, Bigi, Boscá, Shumar, Bacho, Recchia, Popescu, Shulga, Tanwie, Lux, Rank, Ni, Brooks, Yakimchyk, Huanxu, Liu, Cavalleri, Häggström, Verkama, Newbould, Gundlach, Brito-Santana, Amaro, Vajipey, Grover, Wang, Kratish, Li, Gopi, Caciolai, de~Witt, Hernández-Cámara, Rodolà, Robins, Williamson, Cheng, Raynor, Qi, Segev, Fan, Martinson, Wang, Hausknecht, Brenner, Mao, Demian, Kassani, Zhang, Avagian, Scipio, Ragoler, Tan, Sims, Plecnik, Kirtland, Bodur, Shinde, Labrador, Adoul, Zekry, Karakoc, Santos, Shamseldeen, Karim, Liakhovitskaia, Resman, Farina, Gonzalez, Maayan, Anderson, Pena, Kelley, Mariji, Pouriamanesh, Wu, Finocchio, Alarab, Cole, Ferreira, Johnson, Safdari, Dai, Arthornthurasuk, McAlister, Moyano, Pronin, Fan, Ramirez-Trinidad, Malysheva, Pottmaier, Taheri, Stepanic, Perry, Askew, Rodríguez, Minissi, Lorena, Iyer, Fasiludeen, Clark, Ducey, Piza, Somrak, Vergo, Qin, Borbás,
  Chu, Lindsey, Jallon, McInnis, Chen, Semler, Gloor, Shah, Carauleanu, Lauer, Đuc Huy, Shahrtash, Duc, Lewark, Brown, Albanie, Weber, Vaz, Clavier, Fan, e~Silva, Long, Lian, Abramovitch, Jiang, Mendoza, Islam, Gonzalez, Mavroudis, Xu, Kumar, Goswami, Bugas, Heydari, Jeanplong, Jansen, Pinto, Apronti, Galal, Ze-An, Singh, Jiang, of~Arc~Xavier, Agarwal, Berkani, Zhang, Du, de~Oliveira~Junior, Malishev, Remy, Hartman, Tarver, Mensah, Loume, Morak, Habibi, Hoback, Cai, Gimenez, Montecillo, Łucki, Campbell, Sharma, Meer, Gul, Gonzalez, Alapont, Hoover, Chhablani, Vargus, Agarwal, Jiang, Patil, Outevsky, Scaria, Maheshwari, Dendane, Shukla, Cartwright, Bogdanov, Mündler, Möller, Arnaboldi, Thaman, Siddiqi, Saxena, Gupta, Fruhauff, Sherman, Vincze, Usawasutsakorn, Ler, Radhakrishnan, Enyekwe, Salauddin, Muzhen, Maksapetyan, Rossbach, Harjadi, Bahaloohoreh, Sparrow, Sidhu, Ali, Bian, Lai, Singer, Uro, Bateman, Sayed, Menshawy, Duclosel, Bezzi, Jain, Aaron, Tiryakioglu, Siddh, Krenek, Shah, Jin, Creighton,
  Peskoff, EL-Wasif, V, Richmond, McGowan, Patwardhan, Sun, Sun, Zubić, Sala, Ebert, Kaddour, Schottdorf, Wang, Petruzella, Meiburg, Medved, ElSheikh, Hebbar, Vaquero, Yang, Poulos, Zouhar, Bogdanik, Zhang, Sanz-Ros, Anugraha, Dai, Nhu, Wang, Demircali, Jia, Zhou, Wu, He, Chandok, Sinha, Luo, Le, Noyé, Perełkiewicz, Pantidis, Qi, Purohit, Parcalabescu, Nguyen, Winata, Ponti, Li, Dhole, Park, Abbondanza, Wang, Nayak, Caetano, Wong, del Rio-Chanona, Kondor, Francois, Chalstrey, Zsambok, Hoyer, Reddish, Hauser, Rodrigo-Ginés, Datta, Shepherd, Kamphuis, Zhang, Kim, Sun, Yao, Dernoncourt, Krishna, Rismanchian, Pu, Pinto, Wang, Shridhar, Overholt, Briia, Nguyen, David, Bartomeu, Pang, Wecker, Xiong, Li, Huber, Jaeger, Maddalena, Lù, Zhang, Beger, Kon, Li, Sanker, Yin, Liang, Zhang, Agrawal, Yifei, Zhang, Cai, Sonmez, Cozianu, Li, Slen, Yu, Park, Sarti, Briański, Stolfo, Nguyen, Zhang, Perlitz, Hernandez-Orallo, Li, Shabani, Juefei-Xu, Dhingra, Zohar, Nguyen, Pondaven, Yilmaz, Zhao, Jin, Jiang, Todoran, Han,
  Kreuer, Rabern, Plassart, Maggetti, Yap, Geirhos, Kean, Wang, Mollaei, Sun, Yin, Wang, Li, Chang, Wei, Bizeul, Wang, Arrais, Mukherjee, Chamorro-Padial, Liu, Qu, Guan, Bouyamourn, Wu, Plomecka, Chen, Tang, Deng, Subramanian, Xi, Chen, Zhang, Ren, Tu, Kim, Chen, Marjanović, Ha, Luczyna, Ma, Shen, Song, Zhang, Wang, Gendron, Xiao, Smucker, Weng, Lee, Ye, Ermon, Lopez-Miguel, Knights, Gitter, Park, Wei, Chen, Pai, Elkhanany, Lin, Siedler, Fang, Mishra, Zsolnai-Fehér, Jiang, Khan, Yuan, Jain, Lin, Peterson, Wang, Malusare, Tang, Gupta, Fosin, Kang, Dworakowska, Matsumoto, Zheng, Sewuster, Villanueva, Rannev, Chernyavsky, Chen, Banik, Racz, Dong, Wang, Bashmal, Gonçalves, Hu, Bar, Bohdal, Patlan, Dhuliawala, Geirhos, Wist, Kansal, Chen, Tire, Yücel, Christof, Singla, Song, Chen, Ge, Ponkshe, Park, Shi, Ma, Mak, Lai, Moulin, Cheng, Zhu, Zhang, Patil, Jha, Men, Wu, Zhang, Vieira, Aji, Chung, Mahfoud, Hoang, Sperzel, Hao, Meding, Xu, Kostakos, Manini, Liu, Toukmaji, Paek, Yu, Demircali, Sun, Dewerpe, Qin,
  Pflugfelder, Bailey, Morris, Heilala, Rosset, Yu, Chen, Yeo, Jain, Yang, Chigurupati, Chernyavsky, Reddy, Venugopalan, Batra, Park, Tran, Maximiano, Zhang, Liang, Shiyu, Xu, Pan, Suresh, Liu, Gulati, Zhang, Turchin, Bartlett, Scotese, Cao, Wu, Karwowski, Scaramuzza, Nattanmai, McKellips, Cheraku, Suhail, Luo, Deng, Luo, Zhang, Jindel, Paek, Halevy, Baranov, Liu, Avadhanam, Zhang, Cheng, Ma, Fu, Do, Lass, Yang, Sunkari, Bharath, Ai, Leung, Agrawal, Zhou, Chen, Kalpathi, Xu, Wang, Xiao, Maung, Lee, Yang, Yue, Zhao, Yoon, Sun, Singh, Luo, Peng, Osbey, Wang, Echeazu, Yang, Wu, Patel, Kulkarni, Sundarapandiyan, Zhang, Le, Nasim, Yalam, Kasamsetty, Samal, Yang, Sun, Shah, Saha, Zhang, Nguyen, Nagumalli, Wang, Zhou, Wu, Luo, Telluri, Yue, Wang, and Hendrycks}]{phan_humanitys_2025}
Phan L, Gatti A, Han Z, et~al (2025) Humanity's last exam. \doi{10.48550/arxiv.2501.14249}, [Preprint]

\bibitem[{Piña-Iturbe et~al.(2024)Piña-Iturbe, Díaz-Gavidia, Álvarez, Barron-Montenegro, Álvarez Espejo, García, Solís, Constenla-Albornoz, Toro, Olivares-Pacheco, Reyes-Jara, Meng, Bell, and Moreno-Switt}]{pina-iturbe_genomic_2024}
Piña-Iturbe A, Díaz-Gavidia C, Álvarez FP, et~al (2024) Genomic characterisation of the population structure and antibiotic resistance of salmonella enterica serovar infantis in chile, 2009–2022. The Lancet Regional Health - Americas 32:100711. \doi{10.1016/j.lana.2024.100711}

\bibitem[{Rein et~al.(2023)Rein, Hou, Stickland, Petty, Pang, Dirani, Michael, and Bowman}]{rein_gpqa_2023}
Rein D, Hou BL, Stickland AC, et~al (2023) {GPQA}: A graduate-level google-proof q\&a benchmark. \doi{10.48550/arxiv.2311.12022}, [Preprint]

\bibitem[{Roman et~al.(2021)Roman, Shahid, Khan, Koubaa, and Yu}]{roman_citation_2021}
Roman M, Shahid A, Khan S, et~al (2021) Citation intent classification using word embedding. {IEEE} Access 9:9982--9995. \doi{10.1109/access.2021.3050547}

\bibitem[{Sayers et~al.(2024)Sayers, Beck, Bolton, Brister, Chan, Connor, Feldgarden, Fine, Funk, Hoffman, Kannan, Kelly, Klimke, Kim, Lathrop, Marchler-Bauer, Murphy, O’Sullivan, Schmieder, Skripchenko, Stine, Thibaud-Nissen, Wang, Ye, Zellers, Schneider, and Pruitt}]{sayers_database_2024}
Sayers EW, Beck J, Bolton EE, et~al (2024) Database resources of the national center for biotechnology information in 2025. Nucleic Acids Research 53:d20--d29. \doi{10.1093/nar/gkae979}

\bibitem[{Singhal et~al.(2023)Singhal, Azizi, Tu, Mahdavi, Wei, Chung, Scales, Tanwani, Cole-Lewis, Pfohl, Payne, Seneviratne, Gamble, Kelly, Babiker, Schärli, Chowdhery, Mansfield, Demner-Fushman, Agüera~y Arcas, Webster, Corrado, Matias, Chou, Gottweis, Tomasev, Liu, Rajkomar, Barral, Semturs, Karthikesalingam, and Natarajan}]{singhal_large_2023}
Singhal K, Azizi S, Tu T, et~al (2023) Large language models encode clinical knowledge. Nature 620(7972):172--180. \doi{10.1038/s41586-023-06291-2}

\bibitem[{Su et~al.(2019)Su, Prasad, Kan, and Sugiyama}]{su_neural_2019}
Su X, Prasad A, Kan MY, et~al (2019) Neural multi-task learning for citation function and provenance. In: 2019 ACM/IEEE Joint Conference on Digital Libraries (JCDL), pp 394--395, \doi{10.1109/JCDL.2019.00122}

\bibitem[{Taniguti et~al.(2015)Taniguti, Schaker, Benevenuto, Peters, Carvalho, Palhares, Quecine, Nunes, Kmit, Wai, Hausner, Aitken, Berkman, Fraser, Moolhuijzen, Coutinho, Creste, Vieira, Kitajima, and Monteiro-Vitorello}]{taniguti_complete_2015}
Taniguti LM, Schaker PDC, Benevenuto J, et~al (2015) Complete genome sequence of sporisorium scitamineum and biotrophic interaction transcriptome with sugarcane. {PLOS} {ONE} 10(6):e0129318. \doi{10.1371/journal.pone.0129318}

\bibitem[{Teufel et~al.(2006)Teufel, Siddharthan, and Tidhar}]{teufel_automatic_2006}
Teufel S, Siddharthan A, Tidhar D (2006) Automatic classification of citation function. In: EMNLP '06: Proceedings of the 2006 Conference on Empirical Methods in Natural Language Processig, pp 103--110, \urlprefix\url{https://dl.acm.org/doi/10.5555/1610075.1610091}

\bibitem[{Tijsse-Klasen et~al.(2013)Tijsse-Klasen, Pandak, Hengeveld, Takumi, Koopmans, and Sprong}]{tijsse-klasen_ability_2013}
Tijsse-Klasen E, Pandak N, Hengeveld P, et~al (2013) Ability to cause erythema migrans differs between borrelia burgdorferi sensu lato isolates. Parasites \& Vectors 6(1):23. \doi{10.1186/1756-3305-6-23}

\bibitem[{Torene(2025)}]{torene_window_2025}
Torene S (2025) Understanding the impact of increasing llm context windows. \urlprefix\url{https://www.meibel.ai/post/understanding-the-impact-of-increasing-llm-context-windows}, retrieved October 20, 2025

\bibitem[{Tsai et~al.(2018)Tsai, Su, Hsieh, Lin, Chang, Lo, Lin, and Wu}]{tsai_gene_2018}
Tsai TL, Su CC, Hsieh CC, et~al (2018) Gene variations in cis-acting elements between the taiwan and prototype strains of porcine epidemic diarrhea virus alter viral gene expression. Genes 9(12):591. \doi{10.3390/genes9120591}

\bibitem[{Tsolis et~al.(2009)Tsolis, Seshadri, Santos, Sangari, Lobo, de~Jong, Ren, Myers, Brinkac, Nelson, {DeBoy}, Angiuoli, Khouri, Dimitrov, Robinson, Mulligan, Walker, Elzer, Hassan, and Paulsen}]{tsolis_genome_2009}
Tsolis RM, Seshadri R, Santos RL, et~al (2009) Genome degradation in brucella ovis corresponds with narrowing of its host range and tissue tropism. {PLoS} {ONE} 4(5):e5519. \doi{10.1371/journal.pone.0005519}

\bibitem[{Val-Calvo et~al.(2021)Val-Calvo, Miguel-Arribas, Abia, Wu, and Meijer}]{val-calvo_pls20_2021}
Val-Calvo J, Miguel-Arribas A, Abia D, et~al (2021) {pLS}20 is the archetype of a new family of conjugative plasmids harboured by bacillus species. {NAR} Genomics and Bioinformatics 3(4):lqab096. \doi{10.1093/nargab/lqab096}

\bibitem[{Vazquez-Anderson et~al.(2017)Vazquez-Anderson, Mihailovic, Baldridge, Reyes, Haning, Cho, Amador, Powell, and Contreras}]{vazquez-anderson_optimization_2017}
Vazquez-Anderson J, Mihailovic MK, Baldridge KC, et~al (2017) Optimization of a novel biophysical model using large scale in vivo antisense hybridization data displays improved prediction capabilities of structurally accessible {RNA} regions. Nucleic Acids Research 45(9):5523--5538. \doi{10.1093/nar/gkx115}

\bibitem[{W3C(2025)}]{w3_t3_2025}
W3C (2025) Technique \uppercase{T}3: Using standard text formatting conventions for headings. \urlprefix\url{https://www.w3.org/WAI/WCAG21/Techniques/text/T3}, retrieved October 20, 2025

\bibitem[{Wang et~al.(2022)Wang, Sun, Chen, Chen, Sang, and Xie}]{wang_pathogenic_2022}
Wang J, Sun S, Chen Y, et~al (2022) Pathogenic and genomic characterisation of a rabbit sourced pasteurella multocida serogroup f isolate s4. {BMC} Veterinary Research 18(1):288. \doi{10.1186/s12917-022-03381-7}

\bibitem[{Wang et~al.(2024)Wang, Ma, Zhang, Ni, Chandra, Guo, Ren, Arulraj, He, Jiang, Li, Ku, Wang, Zhuang, Fan, Yue, and Chen}]{wang_mmlu-pro_2024}
Wang Y, Ma X, Zhang G, et~al (2024) {MMLU}-pro: A more robust and challenging multi-task language understanding benchmark. \doi{10.48550/arxiv.2406.01574}, [Preprint]

\bibitem[{White et~al.(2021)White, Legione, Taylor-Brown, Fernandez, Higgins, Timms, and Jelocnik}]{white_completing_2021}
White RT, Legione AR, Taylor-Brown A, et~al (2021) Completing the genome sequence of chlamydia pecorum strains {MC}/{MarsBar} and {DBDeUG}: New insights into this enigmatic koala (phascolarctos cinereus) pathogen. Pathogens 10(12):1543. \doi{10.3390/pathogens10121543}

\bibitem[{Wohlers et~al.(2024)Wohlers, Zentgraf, van~der Sande, and Holtmann}]{wohlers_metabolic_2024}
Wohlers H, Zentgraf L, van~der Sande L, et~al (2024) Metabolic engineering of shewanella oneidensis to produce glutamate and itaconic acid. Applied Microbiology and Biotechnology 108(1):36. \doi{10.1007/s00253-023-12879-5}

\bibitem[{Wu et~al.(2012)Wu, Wang, Xu, Liu, Yu, Lou, Zhu, He, Ben, Hu, Götz, and Qu}]{wu_two-component_2012}
Wu Y, Wang J, Xu T, et~al (2012) The two-component signal transduction system {ArlRS} regulates staphylococcus epidermidis biofilm formation in an ica-dependent manner. {PLoS} {ONE} 7(7):e40041. \doi{10.1371/journal.pone.0040041}

\bibitem[{Xie et~al.(2024)Xie, Han, Chen, Xiang, Zhang, He, Xiao, Li, Dai, Feng, Xu, Kang, Kuang, Yuan, Yang, Luo, Zhang, Liu, Xiong, Deng, Jiang, Yao, Li, Yu, Hu, Huang, Liu, Lopez-Lira, Wang, Lai, Wang, Peng, Ananiadou, and Huang}]{xie_finben_2024}
Xie Q, Han W, Chen Z, et~al (2024) {FinBen}: A holistic financial benchmark for large language models. \doi{10.48550/arxiv.2402.12659}, [Preprint]

\bibitem[{Xiong et~al.(2019)Xiong, Liu, Chen, Xiao, Zhu, Gao, Zhang, Chen, Luo, Deng, Chen, Wang, and Chen}]{xiong_new_2019}
Xiong L, Liu S, Chen S, et~al (2019) A new type of {DNA} phosphorothioation-based antiviral system in archaea. Nature Communications 10(1):1688. \doi{10.1038/s41467-019-09390-9}

\bibitem[{Zhang et~al.(2022)Zhang, Zhao, Wang, Chen, Mahmood, Zaib, Zhang, and Sheng}]{zhang_towards_2022}
Zhang Y, Zhao R, Wang Y, et~al (2022) Towards employing native information in citation function classification. Scientometrics 127(11):6557--6577. \doi{10.1007/s11192-021-04242-0}

\bibitem[{Zhang et~al.(2023)Zhang, Wang, Sheng, Mahmood, Zhang, and Zhao}]{zhang_hybrid_2023}
Zhang Y, Wang Y, Sheng QZ, et~al (2023) Hybrid data augmentation for citation function classification. In: 2023 International Joint Conference on Neural Networks ({IJCNN}), pp 1--8, \doi{10.1109/ijcnn54540.2023.10191695}

\bibitem[{Zhang et~al.(2025)Zhang, Wang, Sheng, Yao, Chen, Wang, Mahmood, Zhang, Zaib, Sagar, and Zhao}]{zhang_deep_2025}
Zhang Y, Wang Y, Sheng QZ, et~al (2025) Deep learning meets bibliometrics: A survey of citation function classification. Journal of Informetrics 19(1):101608. \doi{10.1016/j.joi.2024.101608}

\end{thebibliography}
\end{document}